\documentclass{article}

\usepackage[final,nonatbib]{neurips_2020}

\usepackage{multirow}            % tabular cells spanning multiple rows
\usepackage{graphicx}            % figures
\usepackage{duckuments}          % sample images
\usepackage{tabularx}
\usepackage{subcaption}
\usepackage{svg}
\usepackage{float}
\usepackage{adjustbox}

\usepackage{algorithm}
\usepackage{algpseudocode}

\usepackage[utf8]{inputenc} % allow utf-8 input
\usepackage[T1]{fontenc}    % use 8-bit T1 fonts
\usepackage{hyperref}       % hyperlinks
\usepackage{url}            % simple URL typesetting
\usepackage{booktabs}       % professional-quality tables
\usepackage{amsfonts}       % blackboard math symbols
\usepackage{nicefrac}       % compact symbols for 1/2, etc.
\usepackage{microtype}      % microtypography
\usepackage{enumitem}       % control the indent of itemize

\usepackage[
     backend=bibtex,
]{biblatex}
\AtEveryBibitem{
    \clearfield{urlyear}
    \clearfield{urlmonth}
    \clearfield{doi}
}
\usepackage{caption}
\usepackage{xcolor}
\definecolor{darkgreen}{rgb}{0.0, 0.5, 0.0} % This defines a darker green

\title{Composing Open-domain Vision with RAG for Ocean Monitoring and Conservation}

\author{%
  Sepand Dyanatkar\thanks{Corresponding author: \texttt{\href{mailto:sepand@ondeck-ai.com}{sepand@ondeck-ai.com}}. Website: \texttt{\href{https://ondeck.fish}{https://ondeck.fish}}}\ , \ Angran Li, \ Alexander Dungate \\
  OnDeck Fisheries AI\\
}

\begin{document}
\maketitle
 
\begin{abstract}
    Climate change's destruction of marine biodiversity is threatening communities and economies around the world which rely on healthy oceans for their livelihoods.
    The challenge of applying computer vision to niche, real-world domains such as ocean conservation lies in the dynamic and diverse environments where traditional top-down learning struggle with long-tailed distributions, generalization, and domain transfer. 
    Scalable species identification for ocean monitoring is particularly difficult due to the need to adapt models to new environments and identify rare or unseen species. 
    To overcome these limitations, we propose leveraging bottom-up, open-domain learning frameworks as a resilient, scalable solution for image and video analysis in marine applications.
    Our preliminary demonstration uses pretrained vision-language models (VLMs) combined with retrieval-augmented generation (RAG) as grounding, leaving the door open for numerous architectural, training and engineering optimizations.
    We validate this approach through a preliminary application in classifying fish from video onboard fishing vessels, demonstrating impressive emergent retrieval and prediction capabilities without domain-specific training or knowledge of the task itself.
\end{abstract}

\section{Introduction}\label{sec:intro}

Reliable and timely data is critical for addressing climate challenges driving global biodiversity collapse, and threatening 3 billion livelihoods that depend on healthy oceans \cite{noauthor_goal_nodate}.
Thousands of cameras are deployed around the world underwater, on drones, on fishing vessels, and via satellite to monitor marine life and inform critical management tasks.
The scale of the data they produce is already impossible for humans to handle. 
In fisheries monitoring alone, thousands of hours of video are generated per boat, making its manual review crushingly slow, not scalable, and unaffordable.~\cite{wing_advancing_2024}. 

Automated processing of visual data brings key insights about marine activity at an unprecedented scale. 
Traditional vision approaches, such as CNNs and basic vision transformers~\cite{krizhevsky_imagenet_2012, dosovitskiy_image_2020}, require extensive retraining for new environments and perform poorly when faced with rare or unseen species. 
These top-down methods struggle with generalization, domain transfer, and handling long-tailed distributions, and are infeasible for scalable ocean applications.
Generalizable species identification stands to revolutionize marine management by providing comprehensive data on marine life distribution and their responses to climate change - vital for management of sustainable fisheries, invasive species, carbon sequestration, and ecosystem health.

To overcome these challenges, we propose leveraging bottom-up, open-domain learning frameworks as a scalable and resilient solution, where knowledge about tuna species or oil spills are connected in a modular way, even after deployment. 
In this proposal, we show one accessible method for grounded, open-domain vision and highlight the promise of vision-language models (VLMs) combined with retrieval-augmented generation (RAG)~\cite{caffagni_wiki-llava_2024, shen_k-lite_2022, fabian_knowledge_2023, marino_krisp_2021, wu_multi-modal_2021}. 
This approach enhances generalization across diverse environments and enables accurate identification of unseen species by integrating external knowledge during inference.

\section{Related Work}\label{sec:related-work} 

Traditional vision methods such as convolutional neural networks (CNNs), single-stage detectors, and vision transformers~\cite{he_deep_2015, redmon_you_2016, dosovitskiy_image_2020} have been the cornerstone of visual recognition. These top-down methods have shown reliable performance in closed-set, uni-domain climate applications, including environmental monitoring~\cite{norouzzadeh_automatically_2018, norouzzadeh_deep_2019, atlas_wild_2023}. However, their effectiveness diminishes in generalized climate applications where data is often non-IID, and the distribution of objects and events is long-tailed or unexpected~\cite{van_horn_inaturalist_2018, kay_fishnet_2021, katija_fathomnet_2022, stevens_bioclip_2024}. % \todo{mention learning is top-down}

Recent advances in bottom-up, multi-modal approaches offer promising solutions to these challenges. Vision-language models (VLMs) based on CLIP and BLIP-2~\cite{radford_learning_2021, li_blip-2_2023} learn representations of the world through contrastive learning and show emergent capabilities like detection, retrieval, question-answering and broader domain transfer. 
Note that this effectively bridges the modality gap between visual and textual data. 
These models excel in open-domain tasks such as visual question-answering (VQA) and retrieval~\cite{marino_ok-vqa_2019, xu_msr-vtt_2016}%\todo{one more major VQA citation}
, which are crucial for climate monitoring and conservation~\cite{miao_new_2024}. 

The integration of grounding and retrieval-augmented generation (RAG) into these models further enhances their performance by allowing them to access and incorporate external knowledge during inference~\cite{shen_k-lite_2022, sarto_retrieval-augmented_2022, caffagni_wiki-llava_2024, jiang_active_2023, guu_realm_2020}. 
For instance, grounded CLIP~\cite{li_grounded_2022} showcases emergent detection and identification capabilities that compete and surpass SOTA methods. 
Interestingly, \textcite{shen_k-lite_2022} empirically indicates that biodiversity-related domains benefit most from transferable vision models injected with external knowledge (e.g. Flowers102 and OxfordPets). 
Grounding is key for open-ended VQA tasks, where external knowledge is required~\cite{marino_ok-vqa_2019}, but can also alleviate difficulties of class imbalance, domain shift and transfer, and interpretability.
These advances make VLMs combined with RAG particularly promising for addressing the complex challenges posed by climate-related monitoring tasks.
% ~\textcite{hu_reveal_2023} encodes world knowledge into a large-scale memory. 

Despite these advancements, there is still limited work that systematically composes, refines, and scales these approaches across diverse real-world applications. 
Our work addresses this gap by exploring the application of these methods in marine conservation, where the ability to generalize and adapt to new environments is critical for effective and scalable species identification and marine monitoring.

\section{Method}\label{sec:methods}

Current vision research has produced powerful methods that can extrapolate and perform new tasks even without explicit training on them. 
While extremely generalizable, they cannot yet produce specialized outputs they have never seen in training, such as object classification to an unseen class. Thus grounding vision is critical as it provides access to external knowledge, and this is our focus for this preliminary work. 

We outline a framework for building generalizable models adaptable to unseen domains and tasks, with a focus on climate impact,
and marine monitoring applications. 
We integrate 
key components including bottom-up learning and grounding with retrieval-augmented generation (RAG), to address the challenges of scalability, adaptability, and robustness in real-world environments. 
For grounding and RAG, we show a minimal setup using similarity search, opening the door for complete grounding and pipelines in future work.
Design enhancements, fine-tuning, and extra tools (such as prompt optimization, multi-query search, and large context) can greatly improve RAG-based approaches (see Appendix~\ref{appendix:discussion-future-work}).

\begin{figure}[!ht]
    \centering
    \includegraphics[width=0.8\linewidth]{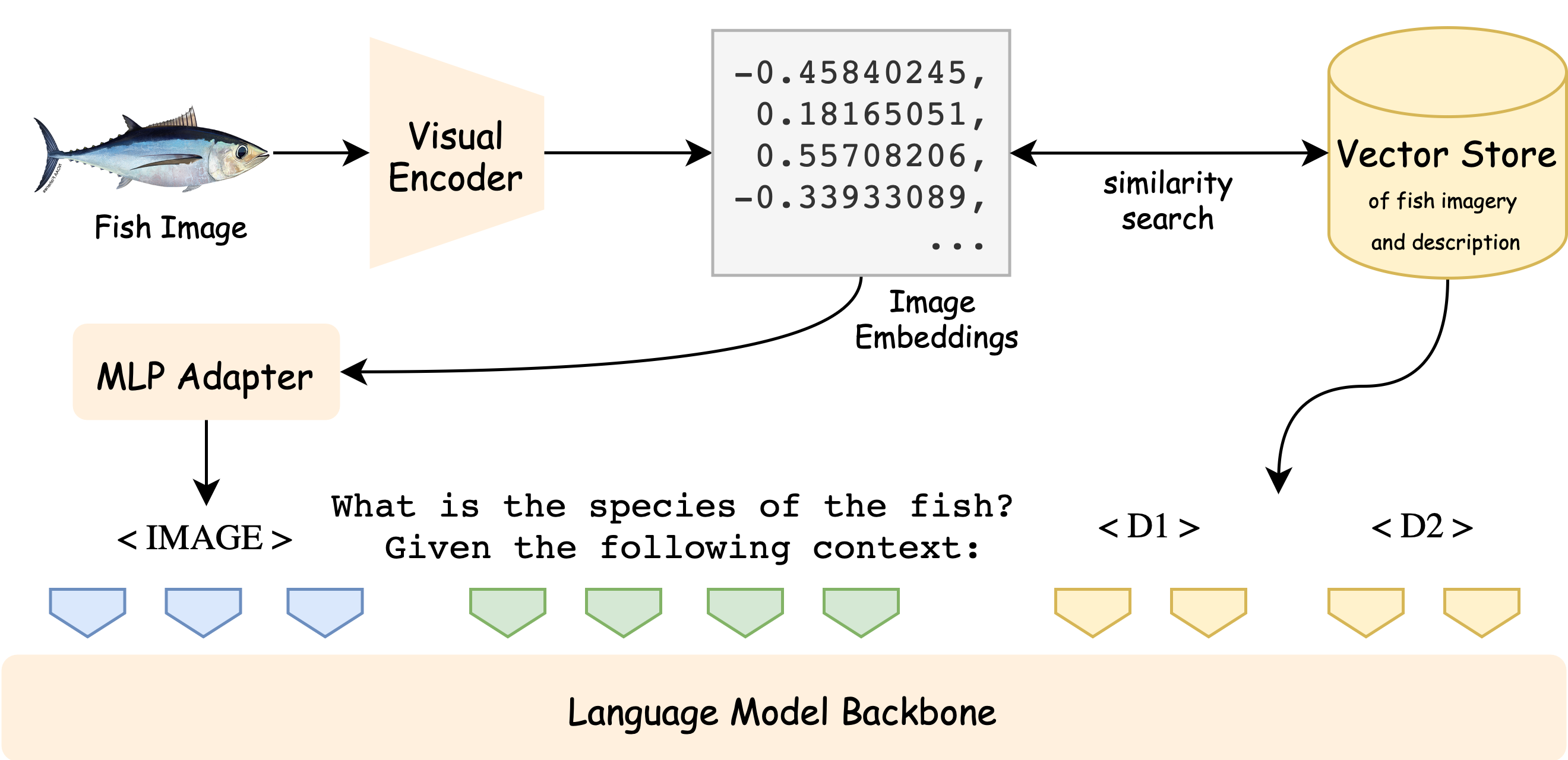}
    \caption{Architecture of visual RAG. The small pentagons with different colours represent  tokens. They are concatenated as input into a language model to generate the final prediction.}
    \label{fig:arch}
\end{figure}

We propose a methodology which uses an open-domain, bottom-up, and task-free model, specifically contrastive learning-style VLMs, combined with RAG to enhance model adaptability and performance across diverse and unseen domains. 
These models can easily transfer between domains, using modular RAG connections for grounding at any time and without model retraining. 
Similar to typical multi-modal retrieval-augmented transformer structures \cite{caffagni_wiki-llava_2024, ramos_smallcap_2023, sarto_retrieval-augmented_2022}, our visual RAG pipeline has three components as shown in Fig.~\ref{fig:arch}:
a CLIP visual encoder to generate image embeddings,
the knowledge base built with image embedding, and 
the backbone where we can evaluate various pre-trained language models. 
As this is preliminary work, we focus on demonstrating the value of information retrieval on visual classification and direct our investigation using an open dataset for fisheries monitoring~\cite{kay_fishnet_2021}.
For a comprehensive version of this method with attention to the entire pipeline, multiple domains, and multiple tasks, see Appendix~\ref{appendix:discussion-future-work} and future work.

\subsection{Image-based Vector Store}
In existing RAG-based vision work, vector stores use text-based keys~\cite{jiang_active_2023, guu_realm_2020, hu_reveal_2023}. Our work is the first to our knowledge to build the knowledge base with an image embedding key. It is motivated by several reasons: a) Poor image quality (Fig.~\ref{fig:input}) causes  direct similarity search between these images' features and text to result in noise and confuses the language model; b) The image embedding database leverages the limited amount of labeled data, helping constrain knowledge retrieval; c) The query itself is static (e.g. object classification), hence adaptive retrieval according to user prompt is unnecessary. 

We produce image embeddings for a small set of reference species using the CLIP encoder and store them as the key in a vector database. The descriptions of different species can then be retrieved by their corresponding image embedding and used as augmented context for answer generation in the final step. Specifically, we produce embeddings of images in the Fishnet validation set using the CLIP encoder and store them as the key in our vector database.

\subsection{Pre-trained Multi-modal LLM}
The LLaVA~\cite{liu_visual_2023} family is widely used as the backbone for multi-modal downstream tasks. We modify the CLIP~\cite{radford_learning_2021} model as visual encoder to generate the query for the similarity search, and use the pre-trained LLaVA 1.5 weights for performance evaluation. To perform a new \textit{set} of tasks (while still) in an emergent manner, VLMs greatly benefit from instruction fine-tuning~\cite{fabian_knowledge_2023}, which can be supported with LoRA\cite{hu_lora_2021}. We leave this for future work (see Appendix~\ref{appendix:discussion-future-work}) building on this proposal. % Fine-tuning with LoRA\cite{hu_lora_2021} can improve alignment, though we did not evaluate it in this preliminary work.

\section{Preliminary Evaluation}
\begin{table}[!tb]
    \centering
    % \fontsize{12}{10}\selectfont
    \captionsetup{belowskip=0pt}
    \caption{Classification accuracy of baseline vs. our VLM-RAG approach on 5 categories. 
    We measure both performance of final prediction (single answer response) and intermediate RAG retrieval. % to verify image encoding and vector DB similarity search
    }
    \begin{tabular}{cccc}
        \toprule
         &  & Accuracy &  \vspace{1mm}\\
        Method & Top-1 & Top-2 & Top-3 \\
        \midrule 
        InceptionV3 (Baseline) & 0.7501 & 0.8312 & 0.9408  \vspace{1mm}\\
        VLM-RAG (Ours, Final Prediction) & 0.8403 & N/A (single answer) & N/A (single answer)\\
        \midrule 
        VLM-RAG (Ours, RAG Retrieval) & \textbf{0.8684} & \textbf{0.9527} & \textbf{0.9781}\\
        \bottomrule
    \end{tabular}
    \vspace{3mm}
    \label{tab:prediction_accuracy}
\end{table}

As was selected for motivation and demonstration in the methods section, we investigate visual RAG on the sample visual task of classification for real-world climate applications. 

We implemented our proposed method using the exact minimal architecture shown of Fig.~\ref{fig:arch}. 
As our dataset, we use \textcite{kay_fishnet_2021}'s Fishnet dataset, version 1.0.0, since it is the largest public dataset of on-deck fisheries activity, and represents a challenge in many of the dimensions discussed in previous sections (data distribution, domain shifts, unseen knowledge).

For visual RAG, we evaluated final prediction performance as this is most comparable to baselines for a classification task (see Table~\ref{tab:prediction_accuracy}). 
We also test visual RAG's intermediate accuracy at the information retrieval step, investigating the performance of the encoder \+ vector search separately. Completing our preliminary ablation, we also tested the pipeline without any grounding (see Fig.~\ref{fig:visualrag-retrieval-precision}).
As our baseline, we used InceptionV3 pretrained on ImageNet with all the same parameters as provided in \textcite{kay_fishnet_2021} using a single NVIDIA A100. 

For intuition on the task difficulty and influence of retrieved descriptions, Fig.~\ref{fig:input} shows typical input image, prompt, and output, using low resolution and partially occluded input, both detrimental for species classification. Without retrieval visually shows our ablation where the category list is explicitly given to the model.

Even with no re-ranking or other optimizations in this preliminary implementation, Table~\ref{tab:prediction_accuracy} shows retrieval performance outperforming both baseline and final prediction. We predict that by advancing the methods outlined in Section~\ref{sec:methods} with steps described in Appendix~\ref{appendix:discussion-future-work} and high-quality embedded descriptions and sample images, the final prediction with outperform RAG retrieval while both see an additional increase. 

In the appendix, we present top-k accuracy of the retrieval process (Fig.~\ref{fig:visualrag-retrieval-topk}), capturing whether the language model has enough information for accurate prediction. 
Figure~\ref{fig:visualrag-retrieval-precision} then show the precision and recall of the final prediction under different RAG settings. 
Additionally, we give a visualisation of our embedding space (Fig.~\ref{fig:embedding_visualization}), to emphasize the task difficulty and the necessity of using visual RAG on on-deck fish images.

\section{Conclusion and Pathways to Climate Impact}\label{sec:conclusion}
Our results demonstrate impressive retrieval and prediction capabilities, without any task or domain-specific training, highlighting the potential of bottom-up learning models to advance scalable marine monitoring. 
Unblocked from the need for expensive domain adaptation, our continued collaboration with fisheries and marine conservation partners will enable faster and more accessible deployments of marine life monitoring for significantly more informed responses to changing climates.

\vspace{3mm}
\begin{minipage}{0.55\textwidth}% adapt widths of minipages to your needs
    \centering
    \includegraphics[width=\linewidth]{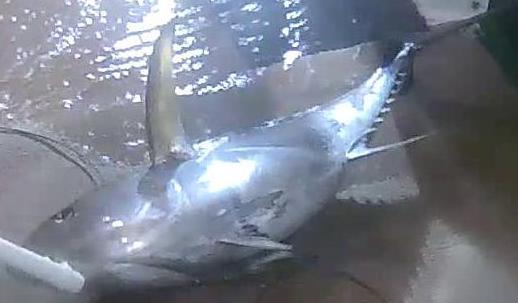}
\end{minipage}%
\hfill
\begin{minipage}{0.39\textwidth}
    \textbf{With retrieval:} \\
    \textit{Q:} What is the species of the fish?\\
    \textit{A:} \textcolor{darkgreen}{The fish in the image is a tuna, as indicated by its torpedo-shaped body, small dorsal and pectoral fins, and metallic blue coloration.}\\
    
    \textbf{Without retrieval, choices provided:} \\
    \textit{Q:} What is the species of the fish?\\
    % It can be either Billfish, Mahi mahi, Opah, Shark, Tuna, or Other (if not sure).\\
    \textit{A:} \textcolor{red}{The fish in the image is a Mahi Mahi}.
\end{minipage}
\captionof{figure}{Example input image and QA with RAG retrieved description (not shown in figure) and without RAG (category list provided but not shown). Images are often low resolution and partly occluded.}\label{fig:input}
\vspace{2mm}

\section*{Acknowledgements}
We would like to thank the entire OnDeck Fisheries AI team for supporting our research and their hard work to translate research into real solutions, accessible and affordable around the world. Thank you to Prof. Graham Taylor's research group in 2023 for conversations about combining non-text modalities with RAG for biodiversity science which encouraged us to conduct our initial literature review and begin this work.

\small
\printbibliography

\normalsize
\clearpage % Force a page break before the appendix

\appendix

\section{Additional Evaluation Details}\label{appendix:results}

\subsection{Precision and Recall by Category w/wo Retrieved Description}\label{appendix-subsec:precision-recall}

\begin{figure}[!htb]
    \centering
    \includegraphics[width=0.8\linewidth]{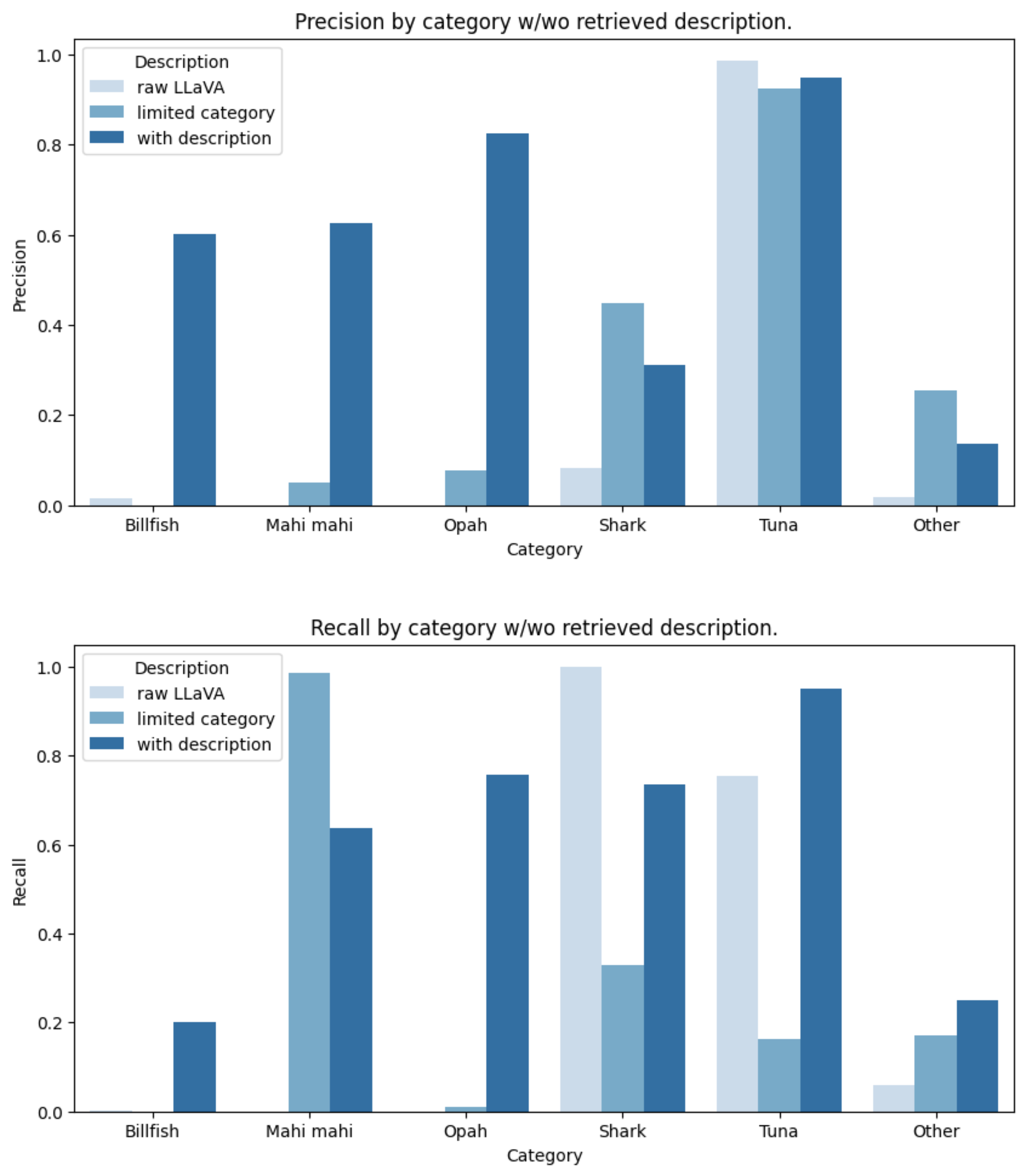}
    \caption{Precision and recall by category in different experiment settings.}
    \label{fig:visualrag-retrieval-precision}
\end{figure}

Figure~\ref{fig:visualrag-retrieval-precision} shows LLaVA final prediction precision and recall by category. The experiment is conducted in three different settings: 

\begin{itemize}[leftmargin=0.4cm]
    \item Raw LLaVA without any categories provided or information retrieved from the database.
    \item LLaVA with category provided and limited to \texttt{Billfish}, \texttt{Mahi mahi}, \texttt{Opah}, \texttt{Shark}, \texttt{Tuna}, and \texttt{Other}.
    \item LLaVA with retrieved description. Here the category limit is implicitly included by description retrieval.
\end{itemize}

\subsection{Top-k Accuracy}
\begin{figure}[!htb]
    \centering
    \includegraphics[width=\linewidth]{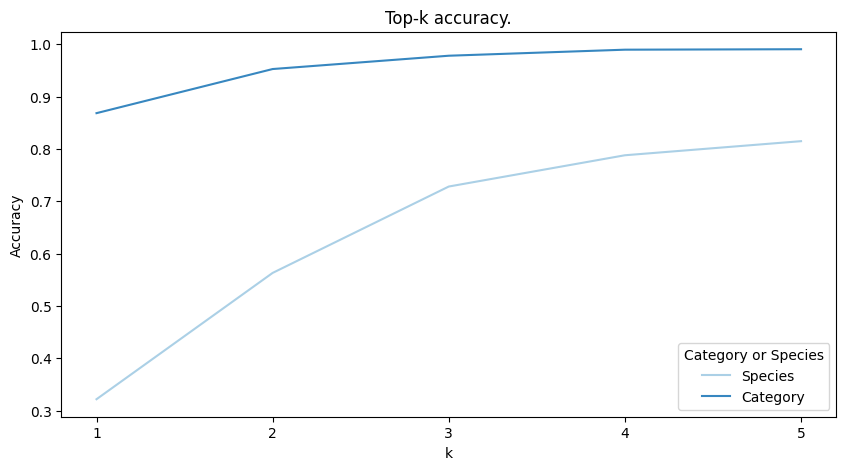}
    \caption{Top-k accuracy for the RAG retrieval process.}
    \label{fig:visualrag-retrieval-topk}
\end{figure}

Figure~\ref{fig:visualrag-retrieval-topk} shows the top-k accuracy of the retrieval process. That is, we are measuring how often the top $k$ retrieved descriptions contain the correct category or species. Higher accuracy of the retrieval should typically mean the model has higher probability to make the correct prediction in the generation step, since it is provided with higher accuracy external knowledge. \newline

The experiments are conducted in two different granularities: category and species, where one category might contain several species that are hard to distinguish. For example, category \texttt{Tuna} contains species \texttt{Albacore}, \texttt{Yellowfin tuna}, \texttt{Skipjack tuna}, \texttt{Bigeye tuna}, and \texttt{Tuna} (which the dataset grouped due to ambiguity).

\subsection{Embedding Space Visualization}
\begin{figure}[!htb]
\minipage{0.49\textwidth}
  \includegraphics[width=\linewidth]{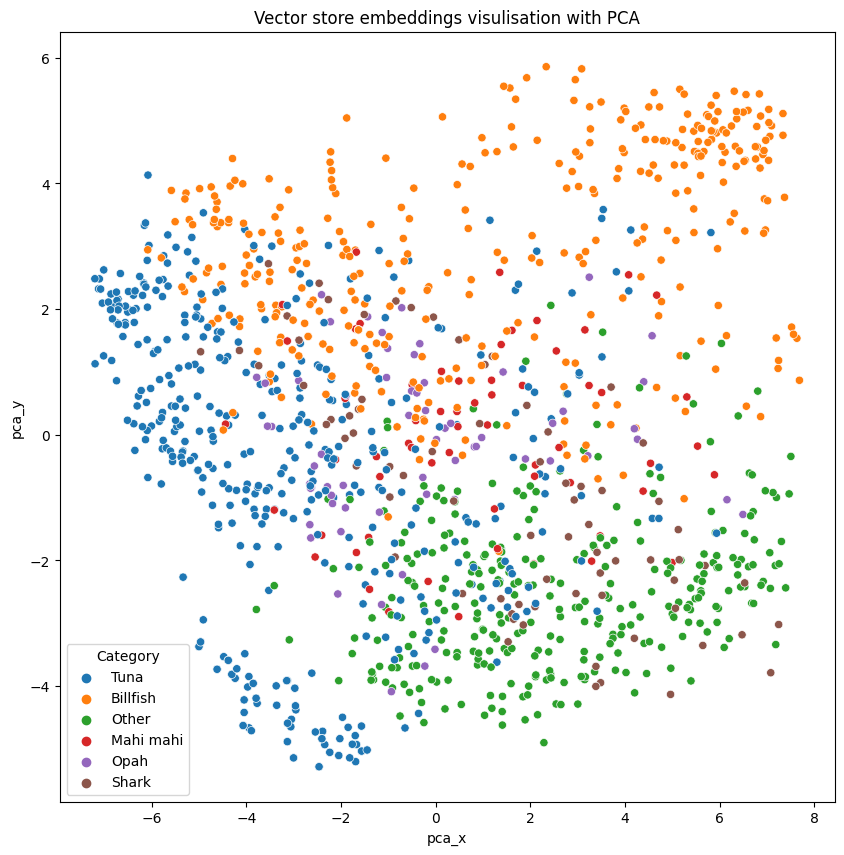}
\endminipage\hfill
\minipage{0.49\textwidth}%
  \includegraphics[width=\linewidth]{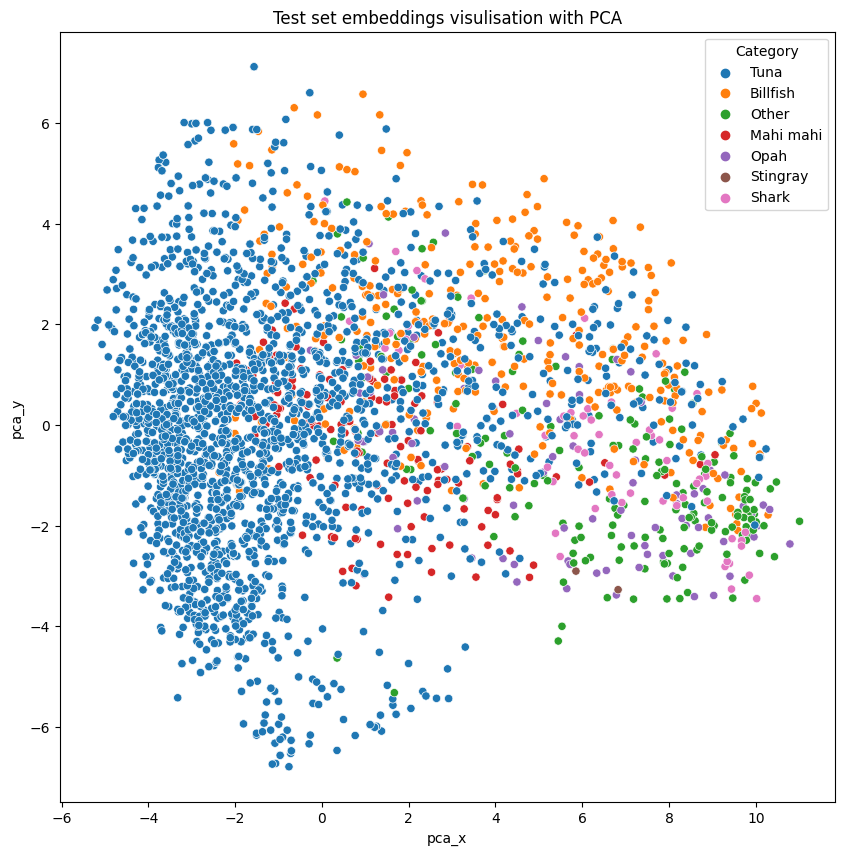}
\endminipage
\caption{Image embedding 2D visualization of vector store and test set.}
\label{fig:embedding_visualization}
\end{figure}

Figure~\ref{fig:embedding_visualization} visualizes the image embedding of our vector store and test set samples, using PCA for dimension reduction. The Figures show:
\begin{itemize}[leftmargin=0.4cm]
    \item The sample feature distribution between two data sets are generally similar to each other. For example, orange samples (\texttt{Billfish}) are in top-right and blue samples (\texttt{Tuna}) are in bottom left.
    \item However, it is still fair to say that there are many differences between the two distributions. For example, the orange samples (\texttt{Billfish}) in the test set are closer to other categories than the corresponding class in the vector store.
    \item The embedding visualization also demonstrates that features from different categories in one data set are fairly mixed up, making it hard to identify the fish species.
\end{itemize}

\section{Discussion and Future Work} \label{appendix:discussion-future-work}
% \todo{Broader interpretation} \ 
% - Where and how do we anticipate it succeeding and why.
The promise of grounded visual reasoning is quite significant. This is even more pronounced with the current momentum towards universal reasoners based on projections of scaling laws for vision models. Grounding these highly generalizable methods can allow us to reach human capacity for analyzing imagery, where a model will look up information it does not have just as a human would. We also emphasize that this work does not explore the "reasoning design", in our case being the choice of (static) pipeline steps and leading into a final VLM-as-predictor with a static query. For example, introducing guided reasoning would allow the architecture to be dynamically adjusted, potentially addressing multi-task capability, and recursive reasoning and attending to harder queries.\newline

Visual reasoning with RAG is generally underexplored. We anticipate and encourage numerous creative and powerful solutions to build on our proposed methodology.\newline

\begin{figure}[!ht]
    \centering
    \includegraphics[width=0.8\linewidth]{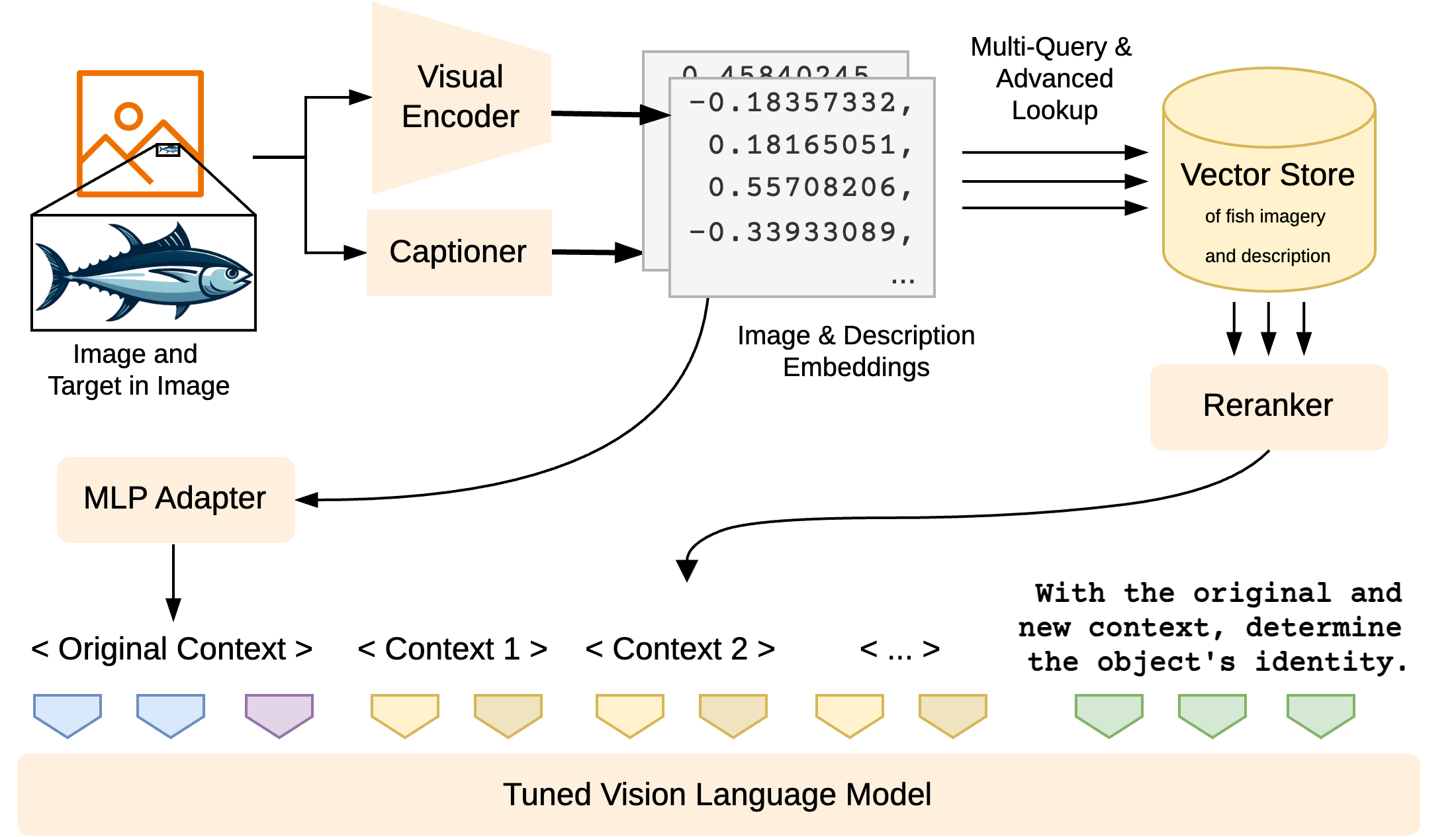}
    \caption{Next iteration of our proposed architecture for visual RAG in classification.}
    \label{fig:arch-next}
\end{figure}

A brief list of dimensions for future work which we anticipate and recommend:
\begin{itemize}
    \item VLM selection and fine-tuning for the task.
    \item RAG-specific improvements (examples in Fig~\ref{fig:arch-next}):
    \begin{itemize}
        \item Multi-query searching
        \item Improved search algorithms
        \item Prompt optimization
        \item Re-ranking results
        \item Optimization of the vector store storage for multiple modalities (going beyond just image and text).
        \item Increased number of results from vector store.
    \end{itemize}
    \item Extracting various streams of information from the imagery (related to multi-query searching). E.g., one stream of nearby background or context, one stream of information about relative sizes of everything in the scene, one stream of just the object in question, etc.
    \item Optimizing for hierarchical step-by-step reasoning, by descending in dimensions of specificity and complexity.
    \item Runtime/In-situ addition \& removal of RAG databases
    \item Extensive evaluation against data with quantifiable and varying human agreement on labels.
\end{itemize}

With all of these enhancements, we also expect a complete ablation to verify the enhancements.

\end{document}